\documentclass{article}
\usepackage[final]{neurips_2019}
\usepackage[utf8]{inputenc}
\usepackage[T1]{fontenc}
\usepackage{hyperref}
\usepackage{url}
\usepackage{booktabs}
\usepackage{amsfonts}
\usepackage{nicefrac}
\usepackage{microtype}
\usepackage{graphicx}

\usepackage{amsmath,amsthm,amssymb}
\usepackage{subcaption}
\usepackage{multirow}
\usepackage{verbatim}

\usepackage{array}
\newcolumntype{L}[1]{>{\raggedright\let\newline\\\arraybackslash\hspace{0pt}}m{#1}}
\newcolumntype{C}[1]{>{\centering\let\\}m{#1}}
\newcolumntype{R}[1]{>{\raggedleft\let\newline\\\arraybackslash\hspace{0pt}}m{#1}}
\usepackage{xcolor}
\usepackage{breqn}

\title{DeepUSPS: Deep Robust Unsupervised Saliency Prediction With Self-Supervision}

\author{
    Duc Tam Nguyen \thanks{Equal contribution, [fixed-term.Maximilian.Dax, Ductam.Nguyen]@de.bosch.com} \thanks{Computer Vision Group,  University of Freiburg, Germany} \thanks{Bosch Research, Bosch GmbH, Germany} , Maximilian Dax \footnotemark[1] \text{}  \footnotemark[3] , Chaithanya Kumar Mummadi \footnotemark[2] \thanks{Bosch Center for AI, Bosch GmbH, Germany}  \\
    \textbf{Thi Phuong Nhung Ngo} \footnotemark[4] , 
    \textbf{Thi Hoai Phuong Nguyen} \thanks{Karlsruhe Institute of Technology, Germany}
    , \textbf{Zhongyu Lou} \footnotemark[3] , \textbf{Thomas Brox} \footnotemark[2]\\
}

\begin{document}

\maketitle

\begin{abstract}
Deep neural network (DNN) based salient object detection in images based on high-quality labels is expensive. Alternative unsupervised approaches rely on careful selection of multiple handcrafted saliency methods to generate noisy pseudo-ground-truth labels.
In this work, we propose a two-stage mechanism for robust unsupervised object saliency prediction, where the first stage involves refinement of the noisy pseudo-labels generated from different handcrafted methods.
Each handcrafted method is substituted by a deep network that learns to generate the pseudo-labels. These labels are refined incrementally in multiple iterations
via our proposed self-supervision technique.
In the second stage, the refined labels produced from multiple networks representing multiple saliency methods are used to train the actual saliency detection network. We show that this self-learning procedure 
outperforms
all the existing unsupervised methods over different datasets. Results are even comparable to those of fully-supervised state-of-the-art approaches. The code is available at \url{https://tinyurl.com/wtlhgo3}.%

 \end{abstract}

\begin{figure}[h]
\centering
     \begin{subfigure}{.18\textwidth}
      \includegraphics[width=1\linewidth]{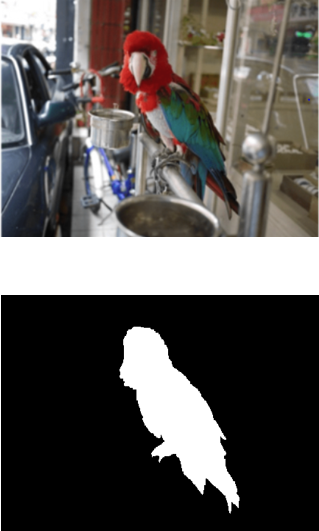}
      \caption{Input and GT}
      \label{fig:input_label}
    \end{subfigure}\hspace{0.04\textwidth}%
    \begin{subfigure}{.38\textwidth}
      \includegraphics[width=0.97\linewidth]{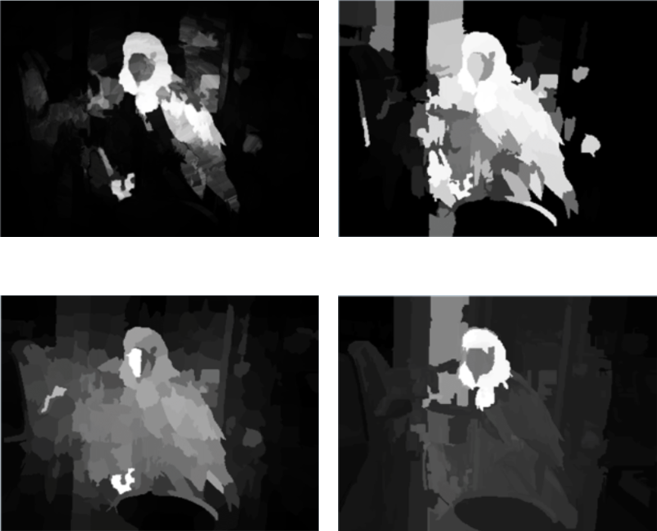}
      \caption{Traditional methods }
      \label{fig:traditional_methods}
    \end{subfigure}\hspace{0.02\textwidth}%
     \begin{subfigure}{.38\textwidth}
      \includegraphics[width=0.95\linewidth]{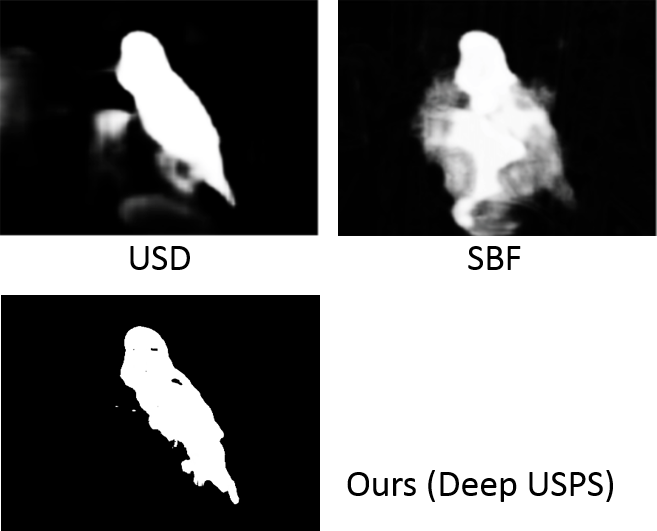}
      \caption{Deep unsupervised methods }
      \label{fig:deep_methods}
    \end{subfigure}%
    
    \caption{Unsupervised object saliency detection based on a given (a) input image. Note that the ground-truth (GT) label is depicted only for illustration purposes and not exploited by any traditional or deep unsupervised methods. 
    (b) Traditional methods use handcrafted priors to predict saliencies and
    (c) deep unsupervised methods  SBF, %
    USD %
    and ours (DeepUSPS) employ the outputs of the handcrafted methods as pseudo-labels in the process of training saliency prediction network. It can be seen that while SBF results in noisy saliency predictions and USD produces smooth saliency maps, our method yields more fine-grained saliency predictions and closely resembles the ground-truth.
    }
    
    \label{fig:qualitative_comaprison}
\end{figure}

\vspace{-5mm}
\section{Introduction}
\vspace{-2mm}

Object saliency prediction aims at finding and segmenting generic objects of interest and help leverage unlabeled information contained in a scene. It can contribute to binary background/foreground segmentation, image caption generation \citep{show2015tell},  semantic segmentation \citep{long2015fully}, or object removal in scene editing \citep{shetty2018adversarial}. In semantic segmentation, for example, the network trained on a fixed set of class labels can only identify objects belonging to these classes, while object saliency detection can highlight an unknown object (e.g., "bear" crossing a street). 
Existing techniques on the saliency prediction task primarily fall under supervised and unsupervised settings. The line of work of supervised approaches \citep{hou2017deeply,luo2017non,zhang2017amulet, zhang2017learning, wang2017stagewise, li2016deepsaliency,wang2016saliency,zhao2015saliency,jiang2013salient,zhu2014saliency} however, requires large-scale clean and pixel-level human-annotated datasets, which are expensive and time-consuming to acquire. Unsupervised saliency methods do not require any human annotations and can work in the wild on arbitrary datasets. These unsupervised methods are further categorized into traditional handcrafted salient object detectors \citep{jiang2013salient,zhu2014saliency,li2013saliency,jiang2013saliency,zou2015harf}  and DNN-based detectors \citep{zhang2018deep, zhang2017supervision}. These traditional methods are based on specific priors, such as center priors \citep{goferman2011context}, global contrast prior \citep{cheng2014global}, and background connectivity assumption \citep{zhu2014saliency}.
Despite their simplicity, these methods perform poorly due to the limited coverage of the hand-picked priors.

\begin{figure}%
\centering
    \includegraphics[width=1\linewidth]{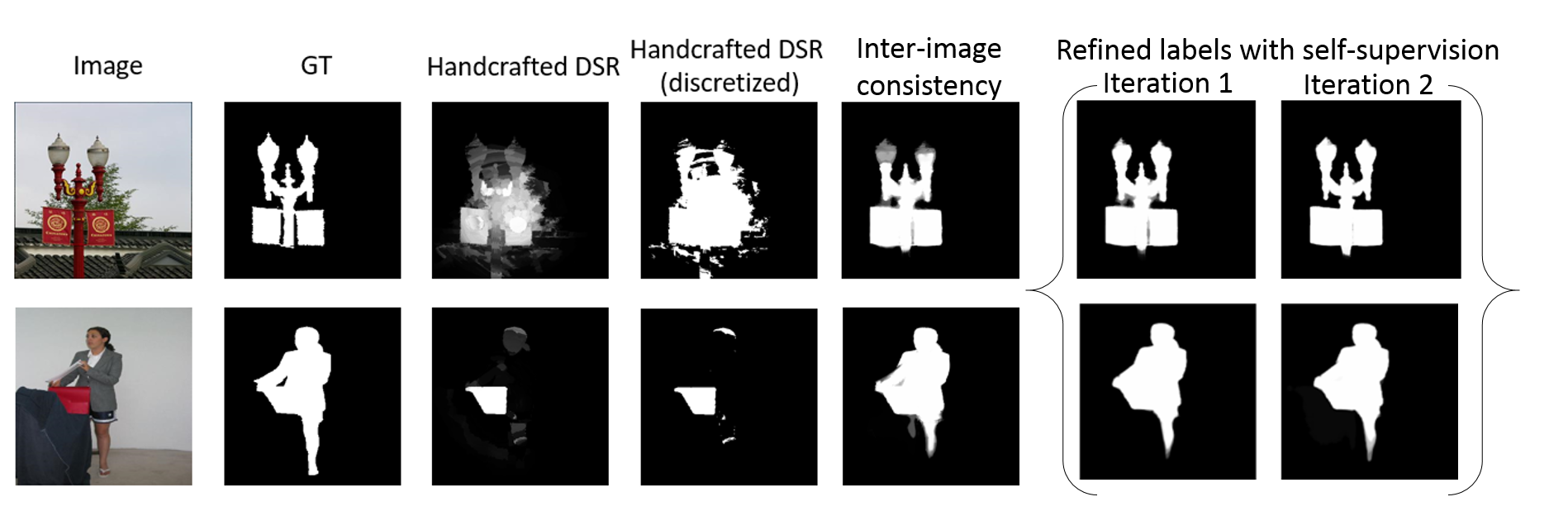}
    \caption{Evolution of refined pseudo-labels from the handcrafted method DSR in our pipeline. Here, we show that the noisy pseudo label from the handcrafted method gets improved with inter-image consistency and further gets refined with our incremental self-supervision technique. While the perceptual differences between pseudo-labels from inter-image consistency and self-supervision technique are minor, we quantitatively show in Table~\ref{tab: ablation study} that this additional refinement improves our prediction results. Results from different handcrafted methods are depicted in Fig. 1 in Appendix.} %
    \label{fig:refined_labels_dsr}
\vspace{-2mm}
\end{figure}

DNN-based approaches leverage the noisy pseudo-label outputs of multiple traditional handcrafted saliency models to provide a supervisory signal for training the saliency prediction network. 
\cite{zhang2017supervision} proposes a method (SBF, 'Supervision by fusion') to fuse multiple saliency models to remove noise from the pseudo-ground-truth labels. 
This method updates the pseudo-labels with the predictions of the saliency detection network and yields very noisy saliency predictions, as shown in Fig.~\ref{fig:deep_methods}. A slightly different approach (USD, 'Deep unsupervised saliency detection') is taken by \cite{zhang2018deep} and introduces an explicit noise modeling module to capture the noise in pseudo-labels of different handcrafted methods. The joint optimization, along with the noise module, enables the learning of the saliency-prediction network to generate the pseudo-noise-free outputs. It does so by fitting different noise estimates on predicted saliency map, based on different noisy pseudo-ground-truth labels.  This method produces smooth predictions of salient objects, as seen in Fig.~\ref{fig:deep_methods} since it employs a noise modeling module to counteract the influence of noise in pseudo-ground-truth labels from handcrafted saliency models.

Both DNN-based methods SBF and USD performs direct pseudo labels fusion on the noisy outputs of handcrafted methods. This implies that the poor-quality pseudo-labels are directly used for training saliency network. Hence, the final performance of the network primarily depends upon the quality of chosen handcrafted methods. On the contrary, a better way is to refine the poor pseudo-labels in isolation in order to maximize the strength of each method. The final pseudo-labels fusion step to train a network should be performed on a set of diverse and high-quality, refined pseudo-labels instead.

More concretely, we propose a systematic curriculum to incrementally refine the pseudo-labels by substituting each handcrafted method with a deep neural network.  The handcrafted methods operate on single image priors and do not infer high-level information such as object shapes and perspective projections. Instead, we learn a function or proxy for the handcrafted saliency method that maps the raw images to pseudo-labels. In other words, we train a deep network to generate the pseudo-labels which benefits from learning the representations across a broad set of training images and thus significantly improve the pseudo-ground-truth labels as seen in Fig.~\ref{fig:refined_labels_dsr} (we refer this effect as inter-images consistency). We further refine our pseudo-labels obtained after the process of inter-image consistency to clear the remaining noise in the labels via the self-supervision technique in an iterative manner. Instead of using pseudo-labels from the handcrafted methods directly as ~\cite{zhang2018deep, zhang2017supervision}, we alleviate the weaknesses of each handcrafted method individually. By doing so, the \emph{diversity} of pseudo-labels from different methods is preserved until the final step when all refined pseudo-labels are fused. The large diversity reduces the over-fitting of the network to the labels noise and results in better generalization capability. %
\begin{figure}[t]
        \begin{center}
            \includegraphics[width=1\linewidth]{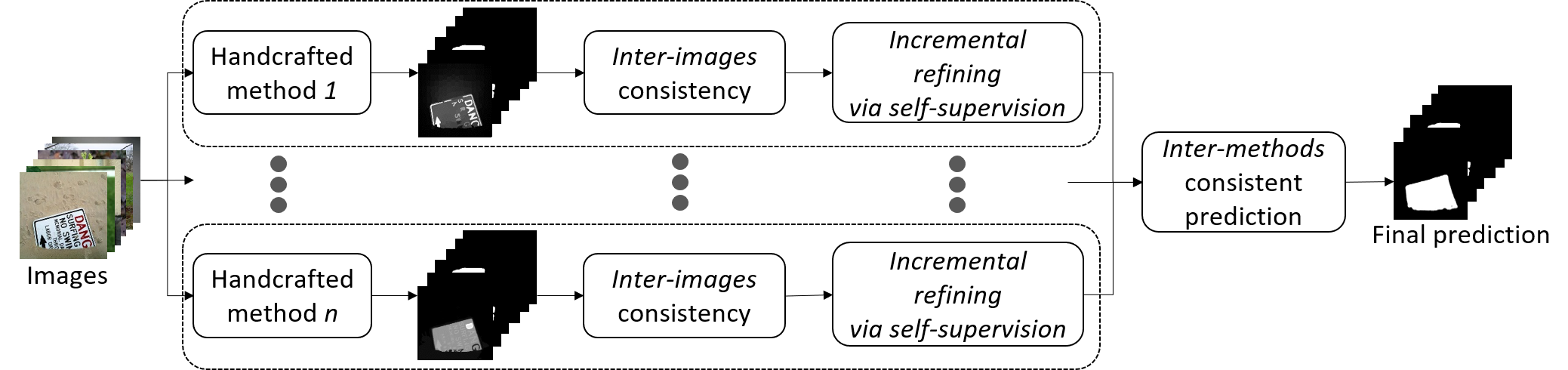}
        \end{center}
        \caption{
        Overview of the sequence of steps involved in our pipeline. Firstly, the training images are processed through different handcrafted methods to generate coarse pseudo-labels. In the second step, which we refer to as \emph{inter-images consistency}, a deep network is learned from the training images and coarse pseudo-labels to generate consistent label outputs, as shown in Fig.~\ref{fig:refined_labels_dsr}. In the next step, the label outputs are further refined with our self-supervision technique in an iterate manner. Lastly, the refined labels from different handcrafted methods are fused for training the saliency prediction network. Details of the individual components in the pipeline are depicted in Fig.~\ref{fig:pipeline detailed}. 
        }
        \label{fig:overview}
        \vspace{-3mm}
\end{figure}

The complete schematic overview of our approach is illustrated in Fig.~\ref{fig:overview}. As seen in the figure, the training images are first processed by different handcrafted methods to create coarse pseudo-labels. In the second step, we train a deep network to predict the pseudo-labels (Fig.~\ref{fig:sfigLv1}) of the corresponding handcrafted method using a image-level loss
to enforce inter-images consistency among the predictions. As seen in Fig.~\ref{fig:refined_labels_dsr}, this step already improves the pseudo-labels over handcrafted methods. In the next step, we employ an iterative self-supervision technique (Fig.~\ref{fig:sfigself-supervision}) that uses historical moving averages (MVA), which acts as an ensemble of various historical models during training (Fig.~\ref{fig:sfigLV1MVA}) to refine the generated pseudo-labels further incrementally. The described pipeline is performed for each handcrafted method individually. In the final step, the saliency prediction network is trained to predict the refined pseudo-labels obtained from multiple saliency methods using a mean image-level loss.\looseness=-1

Our contribution in this work is outlined as follows: we propose a novel systematic mechanism to refine the pseudo-ground-truth labels of handcrafted unsupervised saliency methods iteratively via self-supervision. Our experiments show that this improved supervisory signal enhances the training process of the saliency prediction network. We show that our approach improves the saliency prediction results, outperforms previous unsupervised methods, and is comparable to supervised methods on multiple datasets. Since we use the refined pseudo-labels, the training behavior of the saliency prediction network largely resembles supervised training. Hence, the network has a more stable training process compared to existing unsupervised learning approaches.
\section{Related work}
Various object saliency methods are summarized in \citet{borji2014salient} and evaluated on different benchmarks \citep{borji2015salient}. In the modern literature, the best performances are achieved by deep supervised methods \citep{hou2017deeply,luo2017non,zhang2017amulet, zhang2017learning, wang2017stagewise, li2016deepsaliency,wang2016saliency,zhao2015saliency,jiang2013salient,zhu2014saliency} which all at least use some form of label information.  
The labels might be human-annotated saliency maps or the class of the object at hand. Compared to these fully- and weakly- supervised methods, our approach does not require any label for training. Our method can hence generalize to new datasets without having access to the labels. 

From the literature of deep unsupervised saliency prediction, both \citet{zhang2018deep, zhang2017supervision} use saliency predictions from handcrafted methods as pseudo-labels to train a deep network. \citet{zhang2018deep} proposes a datapoint-dependent noise module to capture the noise among different saliency methods. This additional noise module induces smooth predictions in desired saliency maps.
\citet{zhang2017supervision} defines a manual fusion strategy to combine the pseudo-labels from handcrafted methods on super-pixel and image levels. The resulting, combined labels are a linear combination of existing pseudo-labels. 
This method updates the pseudo-labels with the predictions of a saliency detection network and yields very noisy saliency predictions. In contrast, we refine the pseudo-labels for each handcrafted method in isolation, and hence the diversity of the pseudo-labels is preserved until the last fusion step.

The idea of using handcrafted methods for a pseudo-labels generation has also been adapted by \cite{makansi2018fusionnet} for optical flow prediction. They introduce an assessment network to predict the pixel-wise error of each handcrafted method. Subsequently, they choose the pixel-wise maps to form the best-unsupervised saliency maps. These maps are used as data augmentation for a new domain. However, the best maps are bounded by the quality of existing noisy-maps from the handcrafted methods.
In contrast to their work,  our method improves individual methods gradually by enforcing inter-images consistency, instead of choosing pseudo-labels from the existing set. Further, their method fuses the original pseudo-labels directly in a single step. On the contrary, our fusion step is performed on the \emph{refined} pseudo-labels in a late-stage to preserve diversity. 

From the robust learning perspective, \citet{nguyen2019self} proposes a robust way to learn from wrongly annotated datasets for classification tasks. These techniques can be combined with our presented method to improve the performance further. These advances also improve one-class-training use cases such as anomaly detection \citep{nguyen2019anomaly}, where the models are typically sensitive to noisy labeled data.

Compared to all previous unsupervised saliency methods, we are the first to improve the saliency maps from handcrafted methods in isolation successfully. %
Furthermore, our proposed incremental refining with self-supervision via historical model averaging is unique among this line of research.

\section{DeepUSPS: Deep Unsupervised saliency prediction via self-supervision}

\begin{figure}[h]
\centering
     \begin{subfigure}{.5\textwidth}
      \includegraphics[width=.95\linewidth]{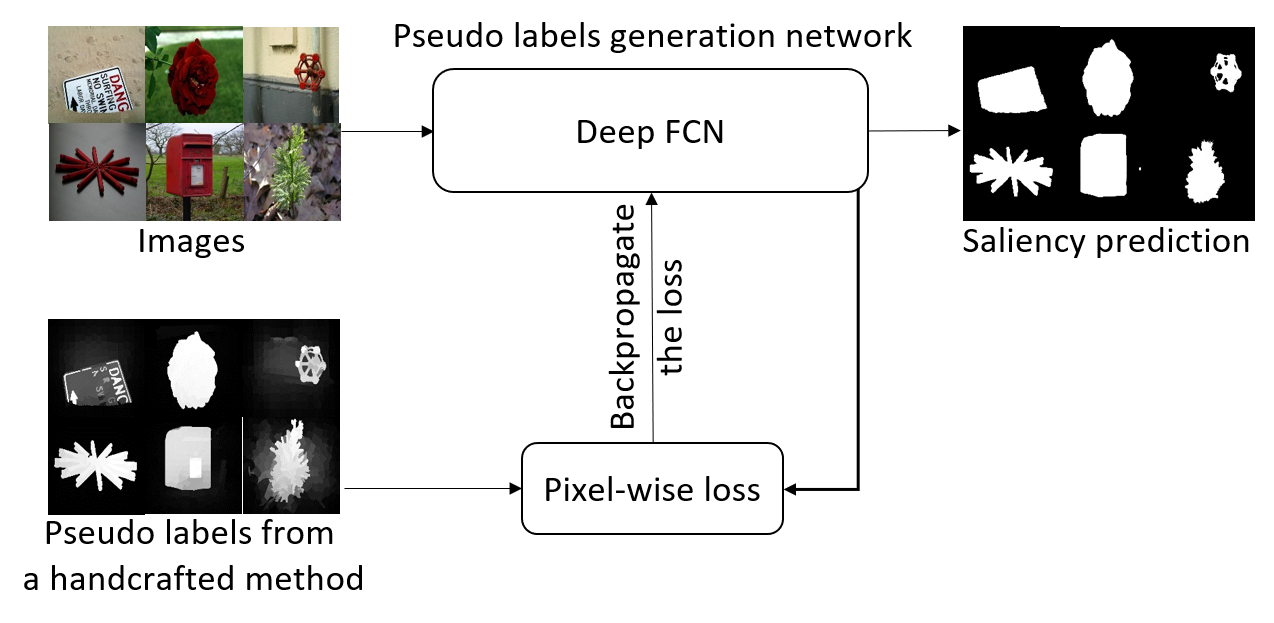}
      \caption{Enforcing inter-images consistency}
      \label{fig:sfigLv1}
    \end{subfigure}%
    \begin{subfigure}{.5\textwidth}
      \includegraphics[width=.95\linewidth]{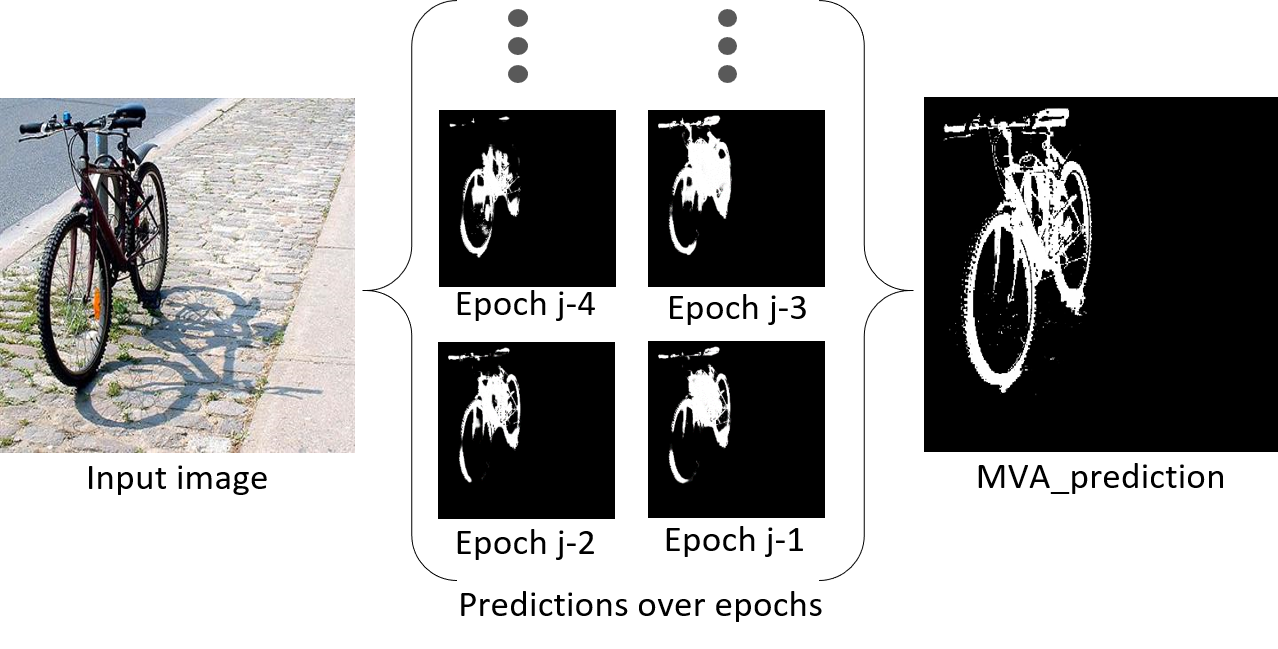}
      \caption{Historical moving averages (MVA) }
      \label{fig:sfigLV1MVA}
    \end{subfigure}%

    \vspace{0.3cm}
    
    \begin{subfigure}{.40\textwidth}
      \includegraphics[width=.95\linewidth]{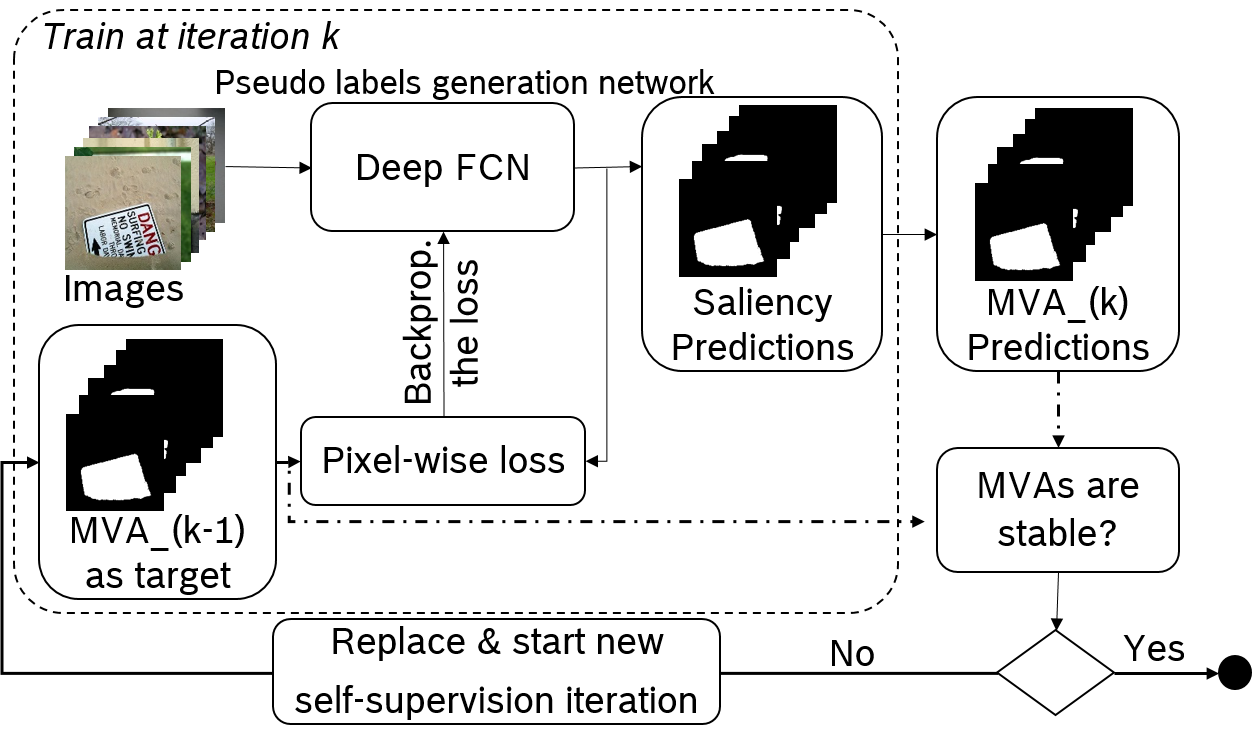}
      \caption{Incremental refining via self-supervision}
      \label{fig:sfigself-supervision}
    \end{subfigure}%
    \begin{subfigure}{.5\textwidth}
      \includegraphics[width=.95\linewidth]{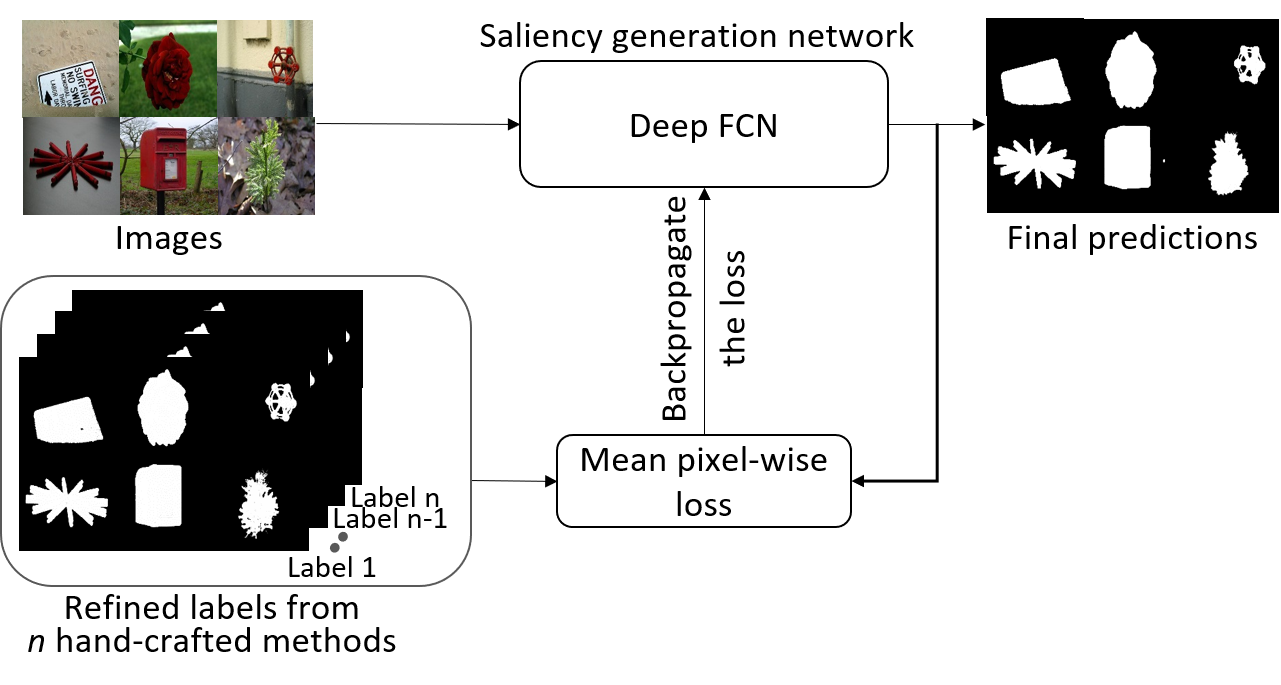}
      \caption{Inter-methods consistent predictions}
      \label{fig:sfig target fusion}
    \end{subfigure}%
    \caption{A detailed demonstration of each step in our pipeline from Fig. \ref{fig:overview}. Handcrafted methods only operate on single images and provide poor-quality pseudo-labels. Hence, (a)-(c) are performed for each handcrafted method separately to refine the pseudo-labels with deep network training. In the final stage (d), the refined pseudo-labels sets are fused by training a network to minimize the averaged loss between different methods.
    }
    \label{fig:pipeline detailed}
    \vspace{-5mm}
\end{figure}

In this section, we explain the technical details of components in the overall pipeline shown in Fig.~\ref{fig:overview}. %
\subsection{Enforcing inter-images consistency with image-level loss }
Handcrafted saliency predictions methods are consistent within an image due to the underlying image priors, but not necessarily consistent across images. They only operate on single image priors and do not infer high-level information such as object shapes and perspective projections. Such inter-images-consistency can be enforced by using outputs from each method as pseudo-labels for training a deep network with an image-level loss. Such a process leads to a refinement of the pseudo-labels suggested by each handcrafted method. \looseness=-1
Let $D$ be the set of training examples and $M$ be a handcrafted method. By $M(x, p)$ we denote the output prediction of method $M$ over pixel $p$ of image $x \in D$. To binarize $M(x, p)$, we use simple function $l(x, p)$ with threshold $\gamma$ such that: $l(x, p) = 1$ if $M(x, p) > \gamma$; $l(x, p) = 0$, otherwise. $\gamma$ equals to $1.5 * \mu_{saliency}$ of the handcrafted method. This discretization scheme counteracts the  method-dependent dynamics in predicting saliency with different degree of uncertainties. The discretization of pseudo-labels make the network less sensitive to over-fitting to the large label noise, compared to fitting to continuous, raw pseudo-labels. \looseness=-1

Given method $M$, let $\theta$ be the set of its corresponding learning parameters in the corresponding FCN and $y(x, p)$ be the output of pixel $p$ in image $x$ respectively. The precision and recall of the prediction over image $x$ w.r.t. The pseudo-labels are straightforward and can be found in the Appendix. The \emph{image-level loss} function w.r.t. each training example $x$ is then defined as $L_\beta= 1 - F_\beta$ where the F-measure $F_\beta$ reflects the weighted harmonic mean of precision and recall such that:
\begin{equation*}
    F_\beta=\left(1+\beta^2\right)\frac{\text{precision} \cdot \text{recall}}{\beta^2~ \text{precision} + \text{recall}}.
\end{equation*}
$L_\beta$ is a linear loss and therefore is more robust to outliers and noise compared to high-order losses such as Mean-Square-Error. The loss is to be minimized by training the FCN for a fixed number of epochs. The fixed number is small to prevent the network from strong over-fitting to the noisy labels. 
\vspace{-3mm}
\paragraph {Historical moving averages of predictions} Due to the large noise ration in the pseudo-labels set, the model snapshots in each training epoch fluctuates strongly. Therefore, a historical moving average of the network saliency predictions $y(x, p)$ is composed during the training procedure, as shown in Fig.~\ref{fig:sfigLV1MVA}.
Concretely, a fully-connected conditional random field (CRF) is applied to  $y(x, p)$ after each forward pass during training. These CRF-outputs are then accumulated into MVA-predictions for each data point at each epoch $k$ as follows:
\begin{equation*}
  \text{MVA}(x, p, k)=
    (1-\alpha) * CRF(y^{j}(x, p)) + \alpha * \text{MVA}(x, p, k-1)
\end{equation*}
Since the MVA is collected during the training process after each forward pass, they do not require additional forward passes for the entire training set. Besides, the predictions are constructed using a large historical model ensemble, where all models snapshots of the training process contribute to the final results. Due to this historical ensembling of saliency predictions, the resulting maps are more robust and fluctuate less strongly compared to taking direct model snapshots. \looseness=-1
\subsection{Incremental pseudo-labels refining via self-supervision}

The moving-average predictions have significantly higher quality than the predictions of the network due to (1) the use of large model ensembles during training and (2) the application of fully connected-CRF. However, the models from the past training iterations in the ensemble are weak due to strong fluctuations, which is a consequence of the training on the noisy pseudo-labels.

To improve the individual models in the ensemble, our approach utilizes the MVA again as the new set of pseudo-labels to train on (Fig.~\ref{fig:sfigself-supervision}). Concretely, the network is reinitialized and minimize the $L_\beta$ again w.r.t. MVA from the last training stage. The process is repeated until the MVA predictions have reached a stable state. By doing so, the diversity in the model ensemble is reduced, but the quality of each model is improved over time. We refer to this process as self-supervised network training with moving average (MVA) predictions.

\subsection{Inter-methods consistent saliency predictions}
Note, that the processes from Fig.~\ref{fig:sfigLv1} to Fig.~\ref{fig:sfigself-supervision} are applied to refine the outputs from each handcrafted method individually. These steps are intended to refine the quality of each method while retaining the underlying designed priors. Furthermore, refining each method in isolation increases the diversity among the pseudo-labels. Hence, the diversity of pseudo-labels is preserved until the final fusion stage. In the last step (Fig.~\ref{fig:sfig target fusion}), the refined saliency maps are fused by minimizing the following loss:
\begin{equation*}
    L_{en} = \frac{1}{n} \Sigma_i L^i_{\beta}
\end{equation*}
where $L^i_{\beta}$ is computed similarly as aforementioned $L_\beta$ using the refined pseudo-labels of method $M_i$; and $\{M_1, \ldots, M_n\}$ are the set of refined handcrafted methods. This fusion scheme is simple and can be exchanged with those from \citet{zhang2018deep, zhang2017supervision,makansi2018fusionnet}.

Our pipeline requires additional computation time to refine the handcrafted methods gradually. Since the training is done in isolation, the added complexity is linear in the number of handcrafted methods. However, the computation of MVAs does not require additional inference steps, since they are accumulated over the training iterations.

\section{Experiments}
We first compare our proposed pipeline to existing benchmarks by following the configuration of \citet{zhang2018deep}. Further, we show in detailed oracle and ablation studies how each component of the pipeline is crucial for the overall competitive performance. Moreover, we analyze the effect of the proposed self-supervision mechanism to improve the label quality over time. \looseness=-1
\subsection{Experiments setup}

Our method is evaluated on traditional object saliency prediction benchmarks \citep{borji2015salient}. Following \citet{zhang2018deep}, we  extract handcrafted maps from MSRA-B  ~\citep{liu2010learning}: 2500 and 500 training and validation images respectively. The remaining test set contains in total 2000 images. Further tests are performed on the ECCSD-dataset ~\citep{yan2013hierarchical} (1000 images), DUT ~\citep{yang2013saliency} (5168 images), SED2 ~\citep{alpert2011image}(100 images).  We re-size all images to 432x432. \looseness=-1

We evaluate the proposed pipeline against different supervised methods, %
traditional unsupervised methods 
and deep unsupervised methods 
from the literature.
We follow the training configuration and setting of the previous unsupervised method \cite{zhang2018deep} to train the saliency detection network.
We use the DRN-network \citep{chen2018deeplab} which is pretrained on CityScapes \citep{cordts2016cityscapes}. The last fully-convolutional layer of the network is replaced to predict a binary saliency mask. Our ablation study also test ResNet101 \citep{he2016deep} that is pretrained on ImageNET ILSVRC \citep{russakovsky2015imagenet}. Our pseudo generation networks are trained %
for
a fixed number of 25 epochs for each handcrafted method and saliency detection network is trained for 200 epochs in the final stage.%
We use ADAM \citep{kingma2014adam} with a momentum of $ 0.9$, batch size 20, a learning rate of 1e-6 in the first step when trained on the handcrafted methods. The learning rate is doubled every time in later self-supervision iteration. Self-supervision is performed for two iterations.  Our models are trained for three times to report the mean and standard deviation. Our proposed pipeline needs about 30 hours of computation time on four Geforce Titan X for training.
\looseness=-1

For the handcrafted methods, we use RBD ('robust background detection') \citep{zhu2014saliency}, DSR ('dense and sparse
reconstruction') \citep{li2013saliency}, MC ('Markov
chain') \citep{jiang2013saliency}, HS ('hierarchy-associated rich features') \citep{zou2015harf}. The $\alpha$-parameter for the exponential moving average for MVA maps is set to $0.7$. Further, the model's predictions are fed into a fully-connected CRF \citep{krahenbuhl2011efficient}. As the evaluation metrics, we utilized Mean-Average-Error (MAE or L1-loss) and weighted F-score with a $\beta^2 = 0.3$ similar to previous works. Furthermore, the analysis of the self-supervision mechanism includes precision and recall that compared against ground-truth-labels. Please refer to Sec. 1 in the Appendix for more details on the definition of these metrics.
\subsection{Evaluation on different datasets}
\begin{figure}[h]
\centering
        \includegraphics[width=0.9\linewidth]{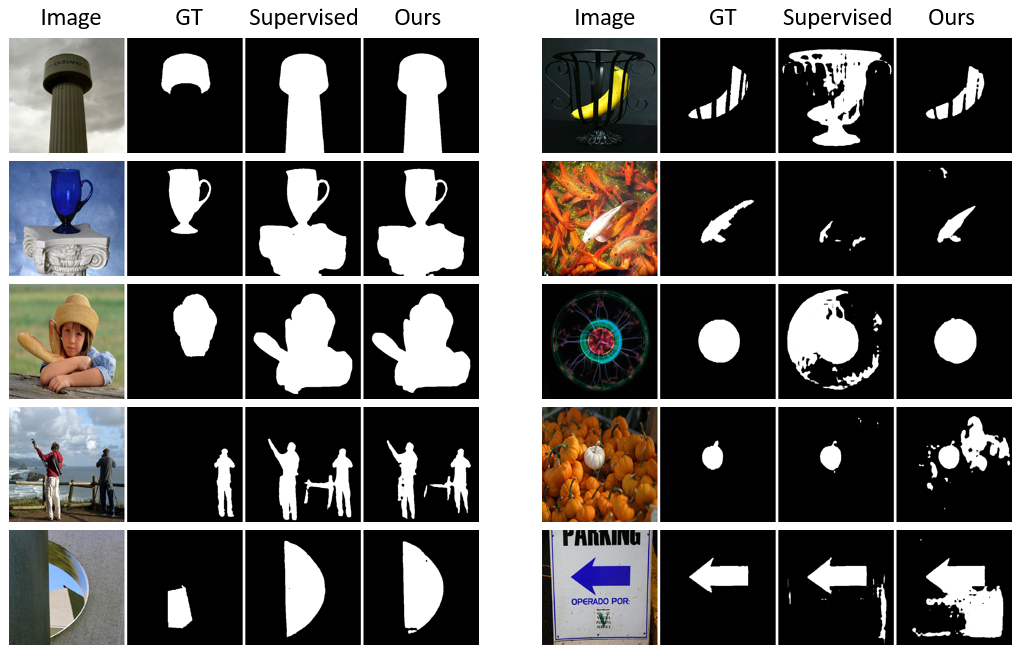}
        \caption{Failure Cases. The left panel shows images (first column) for which both, our approach (fourth column) and the supervised baseline (third column), fail to predict the GT label (second column). In each of these cases, both predictions are close to each other and visually look like justifiable saliency masks despite being significantly different than GT. We found that these kinds of images are indeed responsible for a major part of the bad scores. The right panel shows images for which our predictions are particularly good compared to the baseline prediction, or vice versa. These images are often disturbed by additional intricate details.
        }
        \label{fig:FailureCases_2}
\end{figure}

Tab.~\ref{tab: all benchmarks} shows the performance of our proposed approach on various traditional benchmarks. Our method outperforms other deep unsupervised works consistently on all datasets by a large margin regarding the MAE. Using the F-score metric, we outperform the state-of-the-art (noise modeling from \citet{zhang2018deep} on three out of four datasets. Across the four datasets, our proposed baseline achieves up to 21\%    and 29\%  error reduction on the F-score and MAE-metric, respectively. The effects of different components are to be analyzed in the subsequent oracle test, ablation study, and detailed improvement analysis with self-supervision. Some failure cases are shown in Fig ~\ref{fig:FailureCases_2}.

\begin{table}
  \caption{Comparing our results against various approaches measured in \% of F-score (higher is better) and \% of MAE (lower is better). Bold entries represent the best values in unsupervised methods.%
  }
  \label{tab: all benchmarks}
  \centering
    \begin{tabular}{l*{9}{p{0.8cm}}}
    \toprule
    &   \multicolumn{2}{c}{MSRA-B} & \multicolumn{2}{c}{ECSSD} & \multicolumn{2}{c}{DUT}  & \multicolumn{2}{c}{SED2}\\ 
    Models & $F$$\uparrow$ & MAE$\downarrow$ & $F$$\uparrow$ & MAE$\downarrow$ & $F$$\uparrow$ & MAE$\downarrow$ & $F$$\uparrow$ & MAE$\downarrow$ \\ 
\midrule
\multicolumn{9}{c}{Deep and Supervised}\\
\midrule

~\citet{hou2017deeply}      & 89.41  & 04.74  & 87.96  & 06.99  & 72.90  & 07.60 & 82.36 & 10.14\\
~\citet{luo2017non}         & 89.70  & 04.78  & 89.08  & 06.55  & 73.60  & 07.96 & -  &  - \\
~\citet{zhang2017amulet}       & -   & -   & 88.25  & 06.07  & 69.32  & 09.76 & 87.45 &  06.29\\
~\citet{zhang2017learning}          & -   & -   & 85.21  & 07.97  & 65.95  & 13.21 & 84.44 & 07.42\\
~\citet{wang2017stagewise}         & 85.06  & 06.65  & 82.60  & 09.22  & 67.22  & 08.46 & 74.47 & 11.64 \\
~\citet{li2016deepsaliency}          & -   & -   & 75.89  & 16.01  & 60.45  & 07.58 & 77.78 & 10.74\\
~\citet{wang2016saliency}        & -   & -   & 84.26  & 09.73  & 69.18  & 09.45 & 76.16 & 11.40\\
~\citet{zhao2015saliency}       & 89.66  & 04.91  & 80.61  & 10.19  & 67.15  & 08.85 & 76.60 & 11.62\\
~\citet{jiang2013salient}          & 77.80  & 10.40  & 80.97  & 10.81  & 67.68  & 09.16 & 76.58 & 11.71\\
~\citet{zhu2014saliency}       & 89.73  & 04.67  & 83.15  & 09.06  & 69.02  & 09.71 & 78.40 & 10.14\\
\midrule
\multicolumn{9}{c}{Unsupervised and handcrafted }\\
\midrule
  ~RBD%
  & 75.08  & 11.71  & 65.18  & 18.32  & 51.00  & 20.11 & 79.39 & 10.96\\
  ~DSR%
  & 72.27  & 12.07  & 63.87  & 17.42  & 55.83  & 13.74 & 70.53 & 14.52\\
 ~MC%
 & 71.65  & 14.41  & 61.14  & 20.37  & 52.89  & 18.63 & 66.19 & 18.48\\
 ~HS%
 & 71.29  & 16.09  & 62.34  & 22.83  & 52.05  & 22.74 & 71.68 & 18.69\\
\midrule
\multicolumn{9}{c}{Deep And Unsupervised} \\ 
\midrule
~SBF%
& -   & -   & 78.70  & 08.50  & 58.30  & 13.50 & -  &  -\\
~USD%
& 87.70  & 05.60  & \textbf{87.83}  & 07.04  & 71.56  & 08.60 & 83.80   & 08.81\\              
DeepUSPS (ours)& \textbf{90.31} &  \textbf{03.96}   & 87.42 &  \textbf{06.32}   & \textbf{73.58} &  \textbf{06.25}   & \textbf{84.46} &  \textbf{06.96} \\  
 $\pm$ & 00.10 & 00.03    & 00.46  & 00.10   & 00.87  & 00.02   & 01.00 & 00.06 \\  %

\midrule
\bottomrule
\end{tabular}
\vspace{-3mm}
\end{table}

\subsection{Oracle test and ablation studies}
Tab.~\ref{tab: ablation study} shows an oracle test and an ablation study when a particular component of the proposed pipeline is removed. In the oracle test, we compare the training on ground-truth and oracle labels fusion in the final step, where we choose the pixel-wise best saliency predictions from the refined pseudo-labels. The performance of the oracle labels fusion is on-par with training on the ground-truth, or even slightly better on MSRA-B and SED2. This experiment indicates that DeepUSPS leads to high-quality pseudo-labels. Despite the simple fusion scheme, DeepUSPS approach is only slightly inferior to the oracle label fusion.  Interchanging the architecture to ResNet101, which is pretrained on ImageNet ILSVRC, results in a similarly strong performance. \looseness=-1

The ablation study shows the importance of the components in the pipeline, namely the inter images consistency training and the self-supervision-step. Training on the pseudo-labels from handcrafted methods directly causes consistently poor performance on all datasets. Gradually improving the particular handcrafted maps with our network already leads to substantial performance improvement. The performance further increases with more iterations of self-supervised training. Leaving out the self-supervision stage also decreases the performance of the overall pipeline. 
\looseness=-1
\vspace{-4mm}

\begin{table}
  \caption{Results on extensive ablation studies analyzing the significance of different components
           in our pipeline using F-score and MAE
           on different datasets.
           Our study includes oracle training on GT, oracle label fusion - best pixel-wise choice among different pseudo label maps, using only the pseudo-labels of a single handcrafted method and also analyzing the influence of self-supervision technique over iterations.
  }
  \label{tab: ablation study}
  \centering
  \begin{tabular}{l*{8}{p{0.54cm}}}
    \toprule
    &   \multicolumn{2}{c}{MSRA-B} & \multicolumn{2}{c}{ECSSD} & \multicolumn{2}{c}{DUT}  & \multicolumn{2}{c}{SED2}\\ 
    Models & $F$$\uparrow$ & MAE$\downarrow$ & $F$$\uparrow$ & MAE$\downarrow$ & $F$$\uparrow$ & MAE$\downarrow$ & $F$$\uparrow$ & MAE$\downarrow$ \\ 
\midrule
DeepUSPS (ours)& \textbf{90.31} &  \textbf{03.96}   & 87.42 &  \textbf{06.32}   & \textbf{73.58} &  \textbf{06.25}   & \textbf{84.46} &  \textbf{06.96} \\ 

DeepUSPS (ours)-Resnet101 & 90.05  & 04.17   & 88.17  & 06.41   & 69.60  & 07.71   & 82.60 & 07.31\\ 

\midrule
(Oracle) train on GT   & 91.00  & 03.37   & 90.32  & 04.54   & 74.17  & 05.46   & 80.57 & 07.19\\ %

(Oracle) Labels fusion using GT   & 91.34  & 03.63   & 88.80 & 05.90    & 74.22 & 05.88   & 82.16   & 07.10\\ %
\midrule
Direct fusion of handcrafted methods    & 84.57 & 06.35    & 74.88  & 11.17   & 65.83 & 08.19   & 78.36  & 09.20 \\
\midrule
\multicolumn{9}{c}{Effect of inter-images consistency training}\\
\midrule

Trained on inter-images cons. RBD-maps & 84.49 & 06.25 & 80.62 & 08.82 & 63.86 & 09.17 & 72.05 & 10.33 \\
Trained on inter-images cons. DSR-maps & 85.01 & 06.37 & 80.93 & 09.28 & 64.57 & 08.24 & 65.88 & 10.71 \\
Trained on inter-images cons. MC-maps & 85.72 & 05.80 & 83.33 & 07.73 & 65.65 & 08.51 & 73.90 & 08.95 \\
Trained on inter-images cons. HS-maps & 85.98 & 05.58 & 84.02 & 07.51 & 66.83 & 07.83 & 71.45 & 08.43 \\

\midrule
\multicolumn{9}{c}{Effect of self-supervision}\\
\midrule
No self-supervision    & 89.52  &  04.25  & 85.74  & 06.93  & 72.81  & 06.49 & 84.00 & 07.05\\
\midrule

Trained on refined RBD-maps after iter. 1 & 87.10 & 05.33 & 83.38 & 08.03 & 68.45 & 07.54 & 74.75 & 09.05 \\
Trained on refined RBD-maps after iter. 2 & 88.08 & 04.96 & 84.99 & 07.51 & 70.95 & 06.94 & 78.37 & 08.11 \\
\midrule
Trained on refined DSR-maps after iter. 1 & 87.11 & 05.62 & 82.77 & 08.68 & 67.52 & 07.55 & 71.40 & 09.41 \\
Trained on refined DSR-maps after iter. 2 & 88.34 & 05.17 & 84.73 & 08.08 & 68.82 & 07.21 & 74.24 & 09.06 \\
\midrule
Trained on refined MC-maps after iter. 1 & 87.53 & 05.22 & 84.94 & 07.58 & 67.82 & 07.33 & 70.72 & 09.48 \\
Trained on refined MC-maps after iter. 2 & 88.53 & 04.85 & 85.74 & 07.29 & 69.52 & 06.92 & 73.00 & 09.22 \\
\midrule
Trained on refined HS-maps after iter. 1 & 88.23 & 04.73 & 86.21 & 06.66 & 71.21 & 06.63 & 76.75 & 07.80 \\
Trained on refined HS-maps after iter. 2 & 89.07 & 04.52 & 86.75 & 06.51 & 71.64 & 06.42 & 78.88 & 07.22 \\

\midrule
\bottomrule
\end{tabular}
\end{table}

\subsection{Analyzing the quality of the pseudo label}
\begin{figure}[h]
\centering
     \begin{subfigure}{.25\textwidth}
      \includegraphics[width=1.00\linewidth]{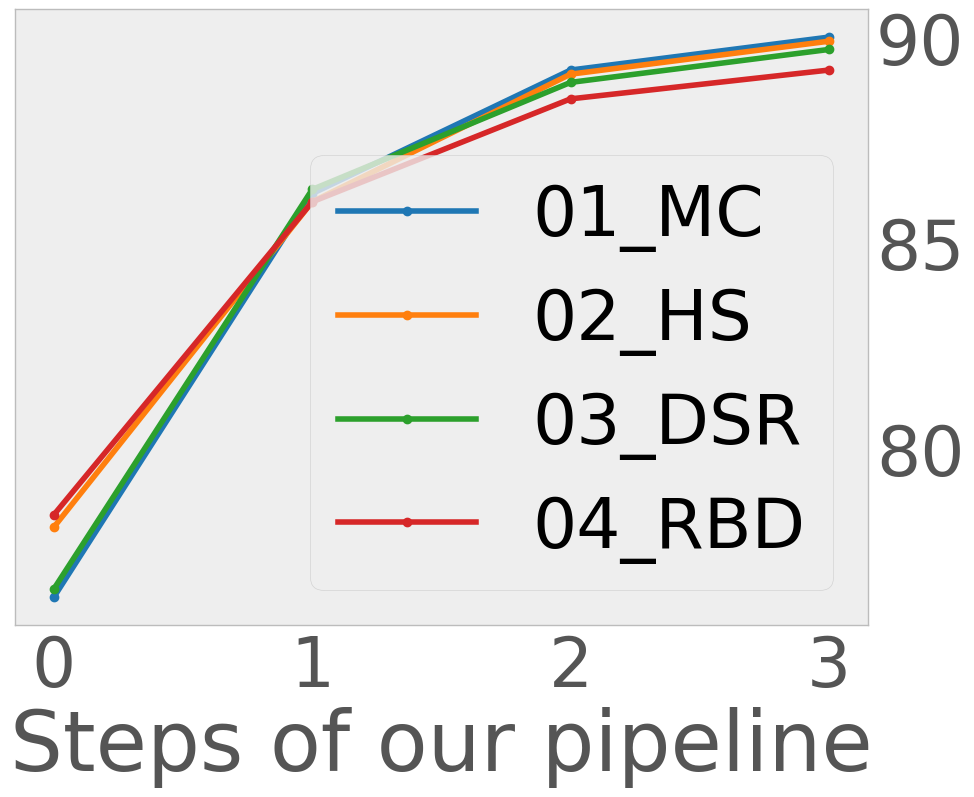}
      \caption{F-score$\uparrow$ saliency maps}
      \label{fig:}
    \end{subfigure}%
    \begin{subfigure}{.25\textwidth}
      \includegraphics[width=1.00\linewidth]{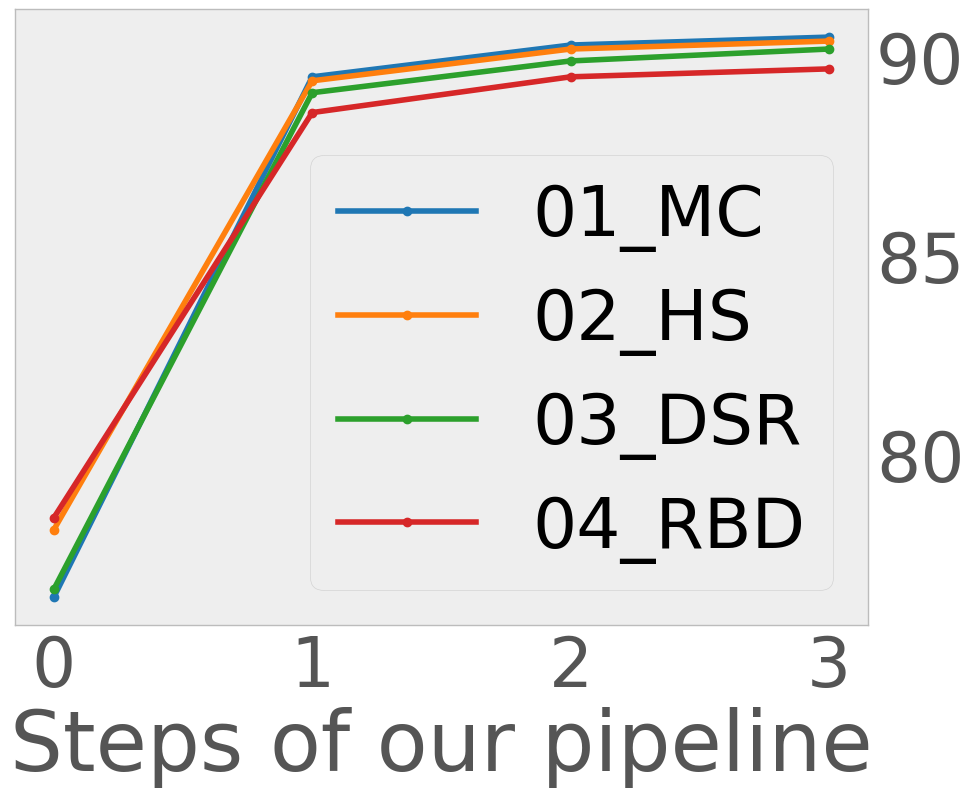}
      \caption{F-score$\uparrow$ MVA maps}
      \label{fig:}
    \end{subfigure}%
     \begin{subfigure}{.25\textwidth}
      \includegraphics[width=1.00\linewidth]{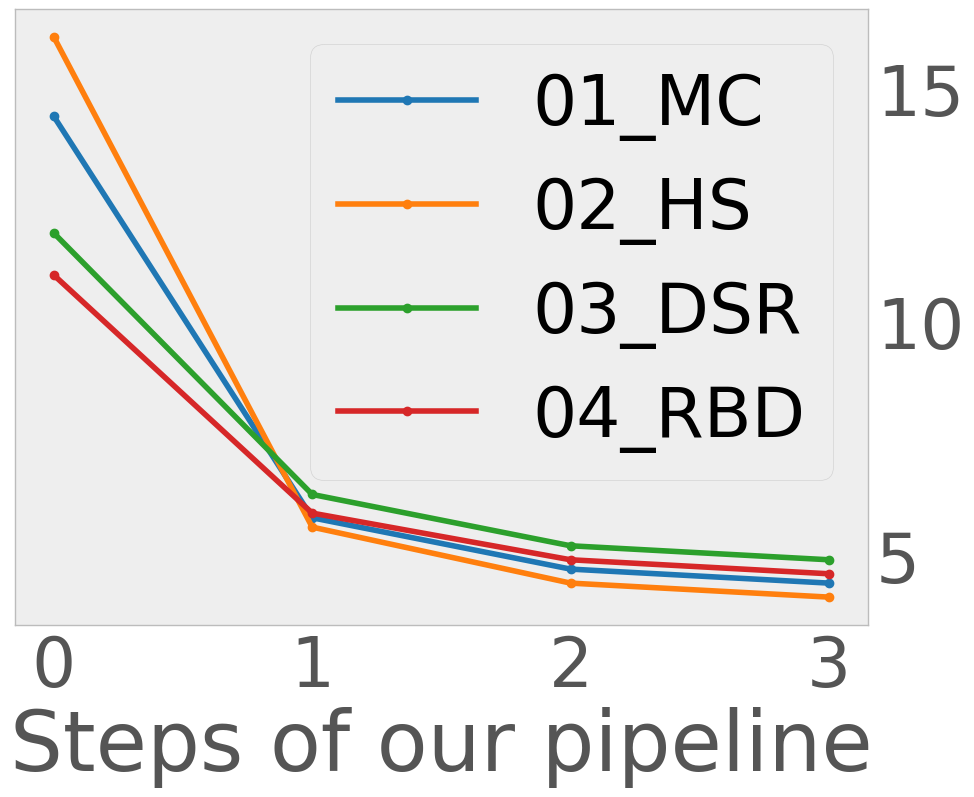}
      \caption{MAE$\downarrow$ saliency maps}
      \label{fig:}
    \end{subfigure}%
    \begin{subfigure}{.25\textwidth}
      \includegraphics[width=1.00\linewidth]{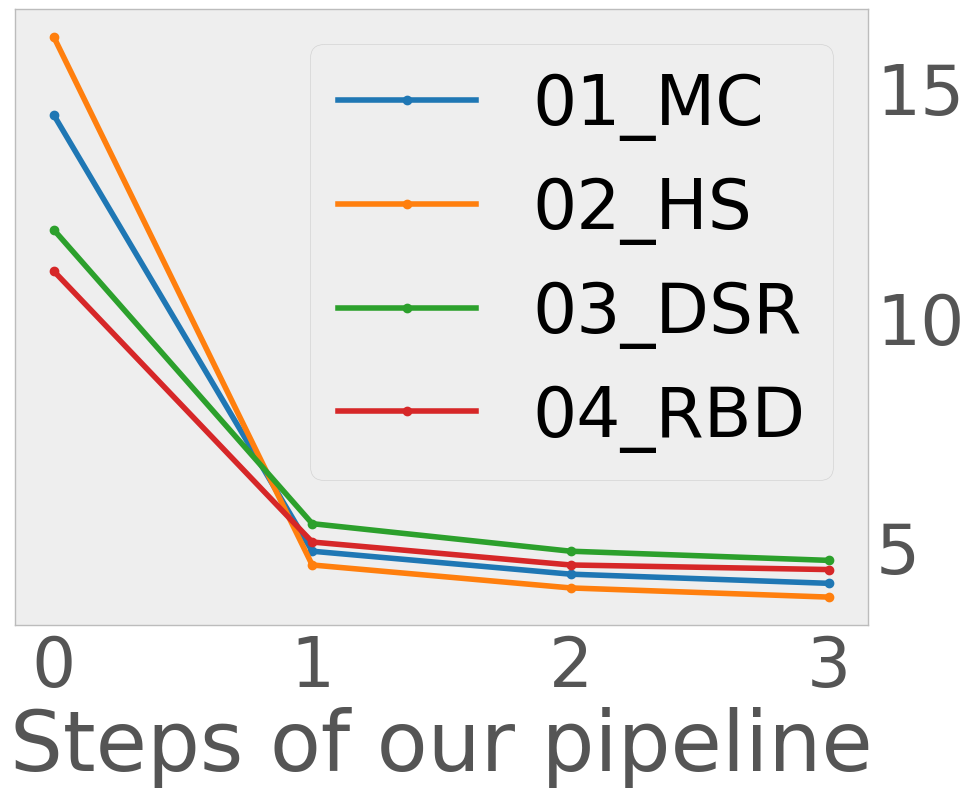}
      \caption{MAE$\downarrow$ MVA-maps}
      \label{fig:}
    \end{subfigure}%
    \caption{Illustrating the improvement of labels quality of predicted saliency maps and aggregated MVA maps on MSRA-B training set from four handcrafted methods over different steps in our pipeline. 
    The steps 0-3 represent measure on the quality of the labels of four different handcrafted methods, inter-images consistency, iteration 1 and iteration 2 of self-supervision with respect to the ground-truth labels. 
    }
    \label{fig:self-supervision all}
    \vspace{-3mm}
\end{figure}

Fig.~\ref{fig:self-supervision all} shows an analysis of the quality of the labels of training images over different steps of our pipeline. 
We analyze the quality of the generated saliency maps (pseudo labels) from the deep networks and also the quality of aggregated MVA maps. %
Here, the quality of the pseudo labels is measured using the ground-truth labels information of the training set.
It can be seen in the figure that the quality of the labels is improved incrementally at each step of our pipeline. Moreover, the quality of MVA maps is shown to be improved rapidly when compared with the saliency maps. Our self-supervision technique further aids in improving the quality of the labels slightly. After few iterations of self-supervision, the F-score and the MAE-score stagnate due to the stable moving-average predictions, and the saliency outputs maps also reach the quality level of the MVA-maps. 
Hence, in the case of offline-testing (when all test data are available at once), the entire proposed procedure might be used to extract high-quality saliency maps. 
In addition, the precision and recall of the quality of the labels are shown in Fig 2 in the Appendix. The handcrafted methods vary strongly in terms of precision as well as recall. This significant variance indicates a large diversity among these pseudo-labels. Our approach is capable of improving the quality of the pseudo labels of each method in isolation.
Thus, the diversity of different methods is preserved until the last fusion step, which enforces inter-methods consistent saliency predictions by the deep network. %

\section{Conclusion}
In this work, we propose to refine the pseudo-labels from different unsupervised handcraft saliency methods in isolation, to improve the supervisory signal for training the saliency detection network. We learn a pseudo-labels generation deep network as a proxy for each handcraft method,  which further enables us to adapt the self-supervision technique to refine the pseudo-labels. We quantitatively show that refining the pseudo-labels iteratively enhances the results of the saliency prediction network and outperforms previous unsupervised techniques by up to 21\% and 29\%  relative error reduction on the F-score and Mean-Average-Error, respectively.  We also show that our results are comparable to the fully-supervised state-of-the-art approaches, which explains that the refined labels are as good as human-annotations. Our studies also reveal that the proposed curriculum learning is crucial to improve the quality of pseudo-labels and hence achieve competitive performance on the object saliency detection tasks. \looseness=-1

\newpage

\bibliographystyle{icml2019}
\bibliography{neurips}

\begin{thebibliography}{33}
\providecommand{\natexlab}[1]{#1}
\providecommand{\url}[1]{\texttt{#1}}
\expandafter\ifx\csname urlstyle\endcsname\relax
  \providecommand{\doi}[1]{doi: #1}\else
  \providecommand{\doi}{doi: \begingroup \urlstyle{rm}\Url}\fi

\bibitem[Alpert et~al.(2011)Alpert, Galun, Brandt, and Basri]{alpert2011image}
Alpert, S., Galun, M., Brandt, A., and Basri, R.
\newblock Image segmentation by probabilistic bottom-up aggregation and cue
  integration.
\newblock \emph{IEEE transactions on pattern analysis and machine
  intelligence}, 34\penalty0 (2):\penalty0 315--327, 2011.

\bibitem[Borji et~al.(2014)Borji, Cheng, Hou, Jiang, and Li]{borji2014salient}
Borji, A., Cheng, M.-M., Hou, Q., Jiang, H., and Li, J.
\newblock Salient object detection: A survey.
\newblock \emph{arXiv preprint arXiv:1411.5878}, 2014.

\bibitem[Borji et~al.(2015)Borji, Cheng, Jiang, and Li]{borji2015salient}
Borji, A., Cheng, M.-M., Jiang, H., and Li, J.
\newblock Salient object detection: A benchmark.
\newblock \emph{IEEE transactions on image processing}, 24\penalty0
  (12):\penalty0 5706--5722, 2015.

\bibitem[Chen et~al.(2018)Chen, Papandreou, Kokkinos, Murphy, and
  Yuille]{chen2018deeplab}
Chen, L.-C., Papandreou, G., Kokkinos, I., Murphy, K., and Yuille, A.~L.
\newblock Deeplab: Semantic image segmentation with deep convolutional nets,
  atrous convolution, and fully connected crfs.
\newblock \emph{IEEE transactions on pattern analysis and machine
  intelligence}, 40\penalty0 (4):\penalty0 834--848, 2018.

\bibitem[Cheng et~al.(2014)Cheng, Mitra, Huang, Torr, and Hu]{cheng2014global}
Cheng, M.-M., Mitra, N.~J., Huang, X., Torr, P.~H., and Hu, S.-M.
\newblock Global contrast based salient region detection.
\newblock \emph{IEEE Transactions on Pattern Analysis and Machine
  Intelligence}, 37\penalty0 (3):\penalty0 569--582, 2014.

\bibitem[Cordts et~al.(2016)Cordts, Omran, Ramos, Rehfeld, Enzweiler, Benenson,
  Franke, Roth, and Schiele]{cordts2016cityscapes}
Cordts, M., Omran, M., Ramos, S., Rehfeld, T., Enzweiler, M., Benenson, R.,
  Franke, U., Roth, S., and Schiele, B.
\newblock The cityscapes dataset for semantic urban scene understanding.
\newblock In \emph{Proceedings of the IEEE conference on computer vision and
  pattern recognition}, pp.\  3213--3223, 2016.

\bibitem[Goferman et~al.(2011)Goferman, Zelnik-Manor, and
  Tal]{goferman2011context}
Goferman, S., Zelnik-Manor, L., and Tal, A.
\newblock Context-aware saliency detection.
\newblock \emph{IEEE transactions on pattern analysis and machine
  intelligence}, 34\penalty0 (10):\penalty0 1915--1926, 2011.

\bibitem[He et~al.(2016)He, Zhang, Ren, and Sun]{he2016deep}
He, K., Zhang, X., Ren, S., and Sun, J.
\newblock Deep residual learning for image recognition.
\newblock In \emph{Proceedings of the IEEE conference on computer vision and
  pattern recognition}, pp.\  770--778, 2016.

\bibitem[Hou et~al.(2017)Hou, Cheng, Hu, Borji, Tu, and Torr]{hou2017deeply}
Hou, Q., Cheng, M.-M., Hu, X., Borji, A., Tu, Z., and Torr, P.~H.
\newblock Deeply supervised salient object detection with short connections.
\newblock In \emph{Proceedings of the IEEE Conference on Computer Vision and
  Pattern Recognition}, pp.\  3203--3212, 2017.

\bibitem[Jiang et~al.(2013{\natexlab{a}})Jiang, Zhang, Lu, Yang, and
  Yang]{jiang2013saliency}
Jiang, B., Zhang, L., Lu, H., Yang, C., and Yang, M.-H.
\newblock Saliency detection via absorbing markov chain.
\newblock In \emph{Proceedings of the IEEE international conference on computer
  vision}, pp.\  1665--1672, 2013{\natexlab{a}}.

\bibitem[Jiang et~al.(2013{\natexlab{b}})Jiang, Wang, Yuan, Wu, Zheng, and
  Li]{jiang2013salient}
Jiang, H., Wang, J., Yuan, Z., Wu, Y., Zheng, N., and Li, S.
\newblock Salient object detection: A discriminative regional feature
  integration approach.
\newblock In \emph{Proceedings of the IEEE conference on computer vision and
  pattern recognition}, pp.\  2083--2090, 2013{\natexlab{b}}.

\bibitem[Kingma \& Ba(2014)Kingma and Ba]{kingma2014adam}
Kingma, D.~P. and Ba, J.
\newblock Adam: A method for stochastic optimization.
\newblock \emph{arXiv preprint arXiv:1412.6980}, 2014.

\bibitem[Kr{\"a}henb{\"u}hl \& Koltun(2011)Kr{\"a}henb{\"u}hl and
  Koltun]{krahenbuhl2011efficient}
Kr{\"a}henb{\"u}hl, P. and Koltun, V.
\newblock Efficient inference in fully connected crfs with gaussian edge
  potentials.
\newblock In \emph{Advances in neural information processing systems}, pp.\
  109--117, 2011.

\bibitem[Li et~al.(2013)Li, Lu, Zhang, Ruan, and Yang]{li2013saliency}
Li, X., Lu, H., Zhang, L., Ruan, X., and Yang, M.-H.
\newblock Saliency detection via dense and sparse reconstruction.
\newblock In \emph{Proceedings of the IEEE International Conference on Computer
  Vision}, pp.\  2976--2983, 2013.

\bibitem[Li et~al.(2016)Li, Zhao, Wei, Yang, Wu, Zhuang, Ling, and
  Wang]{li2016deepsaliency}
Li, X., Zhao, L., Wei, L., Yang, M.-H., Wu, F., Zhuang, Y., Ling, H., and Wang,
  J.
\newblock Deepsaliency: Multi-task deep neural network model for salient object
  detection.
\newblock \emph{IEEE Transactions on Image Processing}, 25\penalty0
  (8):\penalty0 3919--3930, 2016.

\bibitem[Liu et~al.(2010)Liu, Yuan, Sun, Wang, Zheng, Tang, and
  Shum]{liu2010learning}
Liu, T., Yuan, Z., Sun, J., Wang, J., Zheng, N., Tang, X., and Shum, H.-Y.
\newblock Learning to detect a salient object.
\newblock \emph{IEEE Transactions on Pattern analysis and machine
  intelligence}, 33\penalty0 (2):\penalty0 353--367, 2010.

\bibitem[Long et~al.(2015)Long, Shelhamer, and Darrell]{long2015fully}
Long, J., Shelhamer, E., and Darrell, T.
\newblock Fully convolutional networks for semantic segmentation.
\newblock In \emph{Proceedings of the IEEE conference on computer vision and
  pattern recognition}, pp.\  3431--3440, 2015.

\bibitem[Luo et~al.(2017)Luo, Mishra, Achkar, Eichel, Li, and
  Jodoin]{luo2017non}
Luo, Z., Mishra, A., Achkar, A., Eichel, J., Li, S., and Jodoin, P.-M.
\newblock Non-local deep features for salient object detection.
\newblock In \emph{Proceedings of the IEEE Conference on Computer Vision and
  Pattern Recognition}, pp.\  6609--6617, 2017.

\bibitem[Makansi et~al.(2018)Makansi, Ilg, and Brox]{makansi2018fusionnet}
Makansi, O., Ilg, E., and Brox, T.
\newblock Fusionnet and augmentedflownet: Selective proxy ground truth for
  training on unlabeled images.
\newblock \emph{arXiv preprint arXiv:1808.06389}, 2018.

\bibitem[Russakovsky et~al.(2015)Russakovsky, Deng, Su, Krause, Satheesh, Ma,
  Huang, Karpathy, Khosla, Bernstein, et~al.]{russakovsky2015imagenet}
Russakovsky, O., Deng, J., Su, H., Krause, J., Satheesh, S., Ma, S., Huang, Z.,
  Karpathy, A., Khosla, A., Bernstein, M., et~al.
\newblock Imagenet large scale visual recognition challenge.
\newblock \emph{International journal of computer vision}, 115\penalty0
  (3):\penalty0 211--252, 2015.

\bibitem[Shetty et~al.(2018)Shetty, Fritz, and Schiele]{shetty2018adversarial}
Shetty, R.~R., Fritz, M., and Schiele, B.
\newblock Adversarial scene editing: Automatic object removal from weak
  supervision.
\newblock In \emph{Advances in Neural Information Processing Systems}, pp.\
  7706--7716, 2018.

\bibitem[Show(2015)]{show2015tell}
Show, A.
\newblock Tell: Neural image caption generation with visual attention.
\newblock \emph{Kelvin Xu et. al.. arXiv Pre-Print}, 23, 2015.

\bibitem[Wang et~al.(2016)Wang, Wang, Lu, Zhang, and Ruan]{wang2016saliency}
Wang, L., Wang, L., Lu, H., Zhang, P., and Ruan, X.
\newblock Saliency detection with recurrent fully convolutional networks.
\newblock In \emph{European conference on computer vision}, pp.\  825--841.
  Springer, 2016.

\bibitem[Wang et~al.(2017)Wang, Borji, Zhang, Zhang, and Lu]{wang2017stagewise}
Wang, T., Borji, A., Zhang, L., Zhang, P., and Lu, H.
\newblock A stagewise refinement model for detecting salient objects in images.
\newblock In \emph{Proceedings of the IEEE International Conference on Computer
  Vision}, pp.\  4019--4028, 2017.

\bibitem[Yan et~al.(2013)Yan, Xu, Shi, and Jia]{yan2013hierarchical}
Yan, Q., Xu, L., Shi, J., and Jia, J.
\newblock Hierarchical saliency detection.
\newblock In \emph{Proceedings of the IEEE Conference on Computer Vision and
  Pattern Recognition}, pp.\  1155--1162, 2013.

\bibitem[Yang et~al.(2013)Yang, Zhang, Lu, Ruan, and Yang]{yang2013saliency}
Yang, C., Zhang, L., Lu, H., Ruan, X., and Yang, M.-H.
\newblock Saliency detection via graph-based manifold ranking.
\newblock In \emph{Proceedings of the IEEE conference on computer vision and
  pattern recognition}, pp.\  3166--3173, 2013.

\bibitem[Zhang et~al.(2017{\natexlab{a}})Zhang, Han, and
  Zhang]{zhang2017supervision}
Zhang, D., Han, J., and Zhang, Y.
\newblock Supervision by fusion: Towards unsupervised learning of deep salient
  object detector.
\newblock In \emph{Proceedings of the IEEE International Conference on Computer
  Vision}, pp.\  4048--4056, 2017{\natexlab{a}}.

\bibitem[Zhang et~al.(2018)Zhang, Zhang, Dai, Harandi, and
  Hartley]{zhang2018deep}
Zhang, J., Zhang, T., Dai, Y., Harandi, M., and Hartley, R.
\newblock Deep unsupervised saliency detection: A multiple noisy labeling
  perspective.
\newblock In \emph{Proceedings of the IEEE Conference on Computer Vision and
  Pattern Recognition}, pp.\  9029--9038, 2018.

\bibitem[Zhang et~al.(2017{\natexlab{b}})Zhang, Wang, Lu, Wang, and
  Ruan]{zhang2017amulet}
Zhang, P., Wang, D., Lu, H., Wang, H., and Ruan, X.
\newblock Amulet: Aggregating multi-level convolutional features for salient
  object detection.
\newblock In \emph{Proceedings of the IEEE International Conference on Computer
  Vision}, pp.\  202--211, 2017{\natexlab{b}}.

\bibitem[Zhang et~al.(2017{\natexlab{c}})Zhang, Wang, Lu, Wang, and
  Yin]{zhang2017learning}
Zhang, P., Wang, D., Lu, H., Wang, H., and Yin, B.
\newblock Learning uncertain convolutional features for accurate saliency
  detection.
\newblock In \emph{Proceedings of the IEEE International Conference on Computer
  Vision}, pp.\  212--221, 2017{\natexlab{c}}.

\bibitem[Zhao et~al.(2015)Zhao, Ouyang, Li, and Wang]{zhao2015saliency}
Zhao, R., Ouyang, W., Li, H., and Wang, X.
\newblock Saliency detection by multi-context deep learning.
\newblock In \emph{Proceedings of the IEEE Conference on Computer Vision and
  Pattern Recognition}, pp.\  1265--1274, 2015.

\bibitem[Zhu et~al.(2014)Zhu, Liang, Wei, and Sun]{zhu2014saliency}
Zhu, W., Liang, S., Wei, Y., and Sun, J.
\newblock Saliency optimization from robust background detection.
\newblock In \emph{Proceedings of the IEEE conference on computer vision and
  pattern recognition}, pp.\  2814--2821, 2014.

\bibitem[Zou \& Komodakis(2015)Zou and Komodakis]{zou2015harf}
Zou, W. and Komodakis, N.
\newblock Harf: Hierarchy-associated rich features for salient object
  detection.
\newblock In \emph{Proceedings of the IEEE international conference on computer
  vision}, pp.\  406--414, 2015.

\end{thebibliography}

\newpage
\appendix

\section{Loss function}
Due to the pseudo labels noise, the standard Cross-Entropy loss makes the model unstable, as it learns from outliers too strongly. Instead, we use the more robust image-level loss function %
\begin{equation}
    L_\beta=1 - F_\beta,
\end{equation}
where the F-measure $F_\beta$ is defined by the weighted harmonic mean of precision and recall according to
\begin{equation}
    F_\beta=\left(1+\beta^2\right)\frac{precision \cdot recall}{\beta^2~ precision + recall}.
\end{equation}
Precision quantifies how many of the predicted salient pixels are indeed pseudo-ground-truth salient, while recall specifies the fraction of the pseudo-ground-truth salient pixels that are also predicted to be salient. This translates to the following formulas,
\begin{equation}
    precision=\frac{TP}{TP + FP}~~~~~~~~~~~~~~recall=\frac{TP}{TP+TN},
\end{equation}
where $TP$, $FP$ and $TN$ refer to True Positives, False Positives and True Negatives respectively.
In case of discretized prediction $p$ and target $t$ they can be calculated as
\begin{equation}
    \begin{split}
        TP&=\sum_i \left(p_i\cdot t_i\right)\\
        FP&=\sum_i \left(p_i\cdot (1-t_i)\right)\\
        TN&=\sum_i \left((1-p_i)\cdot t_i\right)\\
    \end{split}
\end{equation}
where the sum extends over all pixels $i$. A straightforward generalization to continuous predictions is achieved by dropping the constraint $p_i\in \{0,1\} \forall i$ and allowing for continuous predictions $p\in [0,1]$ instead. The targets remain discrete. This way, the F-measure and hence the loss is differentiable with respect to $p_i$ and can therefore be used for backpropagation.

\newpage
\section{Samples of refined labels}
\begin{figure}[h]
\centering
     \begin{subfigure}{.168\textwidth}
      \includegraphics[width=1\linewidth]{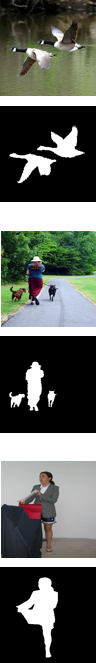}
      \caption{Input and label}
      \label{fig:input_label}
    \end{subfigure}\hspace{0.04\textwidth}%
    \begin{subfigure}{.38\textwidth}
      \includegraphics[width=0.95\linewidth]{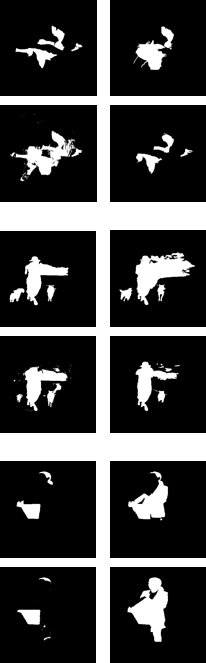}
      \caption{Discretized maps from trad. methods}
      \label{fig:traditional_methods}
    \end{subfigure}\hspace{0.02\textwidth}%
     \begin{subfigure}{.38\textwidth}
      \includegraphics[width=0.95\linewidth]{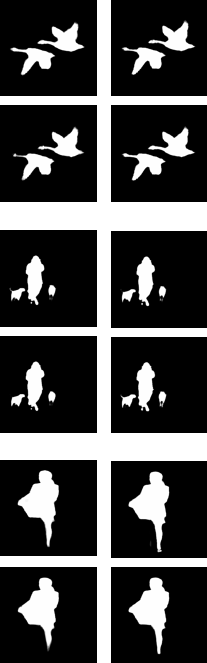}
      \caption{Refined saliency maps}
      \label{fig:deep_methods}
    \end{subfigure}%
    
    \caption{Illustration of pseudo-labels that are generated in our pipeline after refining the coarse pseudo labels of four different traditional methods (MC, HS, RBD, and DSR presented in clockwise order starting from top-left). (a) shows the input image from the training set, (b) depicts the discretized pseudo labels of each handcrafted method and (c) shows the refined pseudo labels after two iterations of self-supervision in our pipeline.}
    
    \label{fig:label_refinement1}
\end{figure}

Figure \ref{fig:label_refinement1} shows several examples of pseudo labels refined in our pipeline.
\vspace{0.5cm}

\newpage
\section{More details of the ablation and oracle studies}
Tab. \ref{tab: ablation study} shows the ablation studies in more details.
\vspace{0.5cm}

\begin{table}[h]
  \caption{Results on extensive ablation studies analyzing the significance of different components in our pipeline for object saliency prediction. We measure the F-score (higher is better) and MAE (lower is better) on four different datasets. Here, oracle tests include the training on ground-truth (GT) labels and label fusion using GT - training on the best pixel-wise choice (measured using GT) among different pseudo label maps (this results in a single pseudo label map). We also analyzed the prediction results of the network that is trained only on pseudo labels of a single handcrafted method. Furthermore, we show the influence of self-supervision technique on the prediction results over iterations. }
  \label{tab: ablation study}
  \centering

\begin{tabular}{l*{8}{p{0.56cm}}}
    \toprule
    &   \multicolumn{2}{c}{MSRA-B} & \multicolumn{2}{c}{ECSSD} & \multicolumn{2}{c}{DUT}  & \multicolumn{2}{c}{SED2}\\ 
    Models & $F$$\uparrow$ & MAE$\downarrow$ & $F$$\uparrow$ & MAE$\downarrow$ & $F$$\uparrow$ & MAE$\downarrow$ & $F$$\uparrow$ & MAE$\downarrow$ \\ 
\midrule

DeepUSPS (ours)& \textbf{90.31} &  \textbf{03.96}   & 87.42 &  \textbf{06.32}   & \textbf{73.58} &  \textbf{06.25}   & \textbf{84.46} &  \textbf{06.96} \\  
$\pm$   & 00.10 & 00.03    & 00.46  & 00.10   & 00.87  & 00.02   & 01.00 & 00.06 \\  %
No CRF & 90.21  & 03.99  & 87.38  & 06.35  & 73.36  & 06.31  & 84.71  & 06.92 \\ %
$\pm$ & 00.12  & 00.03  & 00.13  & 00.04  & 00.21  & 00.08  & 00.45  & 00.08\\  %

\midrule
(Oracle) train on GT   & 91.00  & 03.37   & 90.32  & 04.54   & 74.17  & 05.46   & 80.57 & 07.19\\ %
$\pm$   & 00.10 & 00.03    & 00.46  & 00.10   & 00.87  & 00.02   & 01.00 & 00.06 \\  %

(Oracle) Labels fusion using GT   & 91.34  & 03.63   & 88.80 & 05.90    & 74.22 & 05.88   & 82.16   & 07.10\\ %
$\pm$   & 00.06 & 00.04   & 00.61 & 00.19   & 00.78  & 00.06   & 01.28  & 00.19 \\  %

\midrule
Direct fusion of handcrafted methods    & 84.57 & 06.35    & 74.88  & 11.17   & 65.83 & 08.19   & 78.36  & 09.20 \\
$\pm$   & 00.07 & 00.01    & 00.37 & 00.08   & 00.16  & 00.05    & 00.28 & 00.12 \\ 

\midrule

\multicolumn{9}{c}{Effect of inter-images consistency training}\\
\midrule

Trained on inter-images cons. RBD-maps & 84.49 & 06.25 & 80.62 & 08.82 & 63.86 & 09.17 & 72.05 & 10.33 \\
Trained on inter-images cons. DSR-maps & 85.01 & 06.37 & 80.93 & 09.28 & 64.57 & 08.24 & 65.88 & 10.71 \\
Trained on inter-images cons. MC-maps & 85.72 & 05.80 & 83.33 & 07.73 & 65.65 & 08.51 & 73.90 & 08.95 \\
Trained on inter-images cons. HS-maps & 85.98 & 05.58 & 84.02 & 07.51 & 66.83 & 07.83 & 71.45 & 08.43 \\

\midrule
\multicolumn{9}{c}{Effect of self-supervision}\\
\midrule
No self-supervision    & 89.52  &  04.25  & 85.74  & 06.93  & 72.81  & 06.49 & 84.00 & 07.05\\
\midrule

Trained on refined RBD-maps after iter. 1 & 87.10 & 05.33 & 83.38 & 08.03 & 68.45 & 07.54 & 74.75 & 09.05 \\
Trained on refined RBD-maps after iter. 2 & 88.08 & 04.96 & 84.99 & 07.51 & 70.95 & 06.94 & 78.37 & 08.11 \\
\midrule
Trained on refined DSR-maps after iter. 1 & 87.11 & 05.62 & 82.77 & 08.68 & 67.52 & 07.55 & 71.40 & 09.41 \\
Trained on refined DSR-maps after iter. 2 & 88.34 & 05.17 & 84.73 & 08.08 & 68.82 & 07.21 & 74.24 & 09.06 \\
\midrule
Trained on refined MC-maps after iter. 1 & 87.53 & 05.22 & 84.94 & 07.58 & 67.82 & 07.33 & 70.72 & 09.48 \\
Trained on refined MC-maps after iter. 2 & 88.53 & 04.85 & 85.74 & 07.29 & 69.52 & 06.92 & 73.00 & 09.22 \\
\midrule
Trained on refined HS-maps after iter. 1 & 88.23 & 04.73 & 86.21 & 06.66 & 71.21 & 06.63 & 76.75 & 07.80 \\
Trained on refined HS-maps after iter. 2 & 89.07 & 04.52 & 86.75 & 06.51 & 71.64 & 06.42 & 78.88 & 07.22 \\

\midrule

\bottomrule
\end{tabular}
\end{table}

\newpage
\section{Self-supervision}
\begin{figure}[h]
\centering
    
     \begin{subfigure}{.50\textwidth}
      \includegraphics[width=1.00\linewidth]{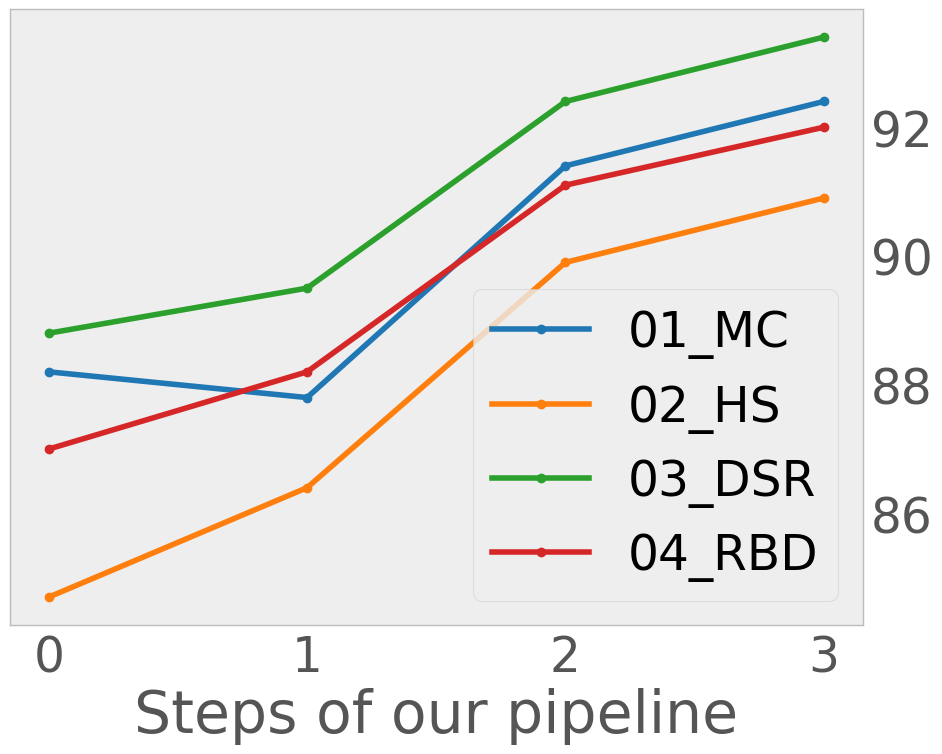}
      \caption{Precision $\uparrow$ of saliency maps}
      \label{fig:}
    \end{subfigure}%
     \begin{subfigure}{.50\textwidth}
      \includegraphics[width=1.00\linewidth]{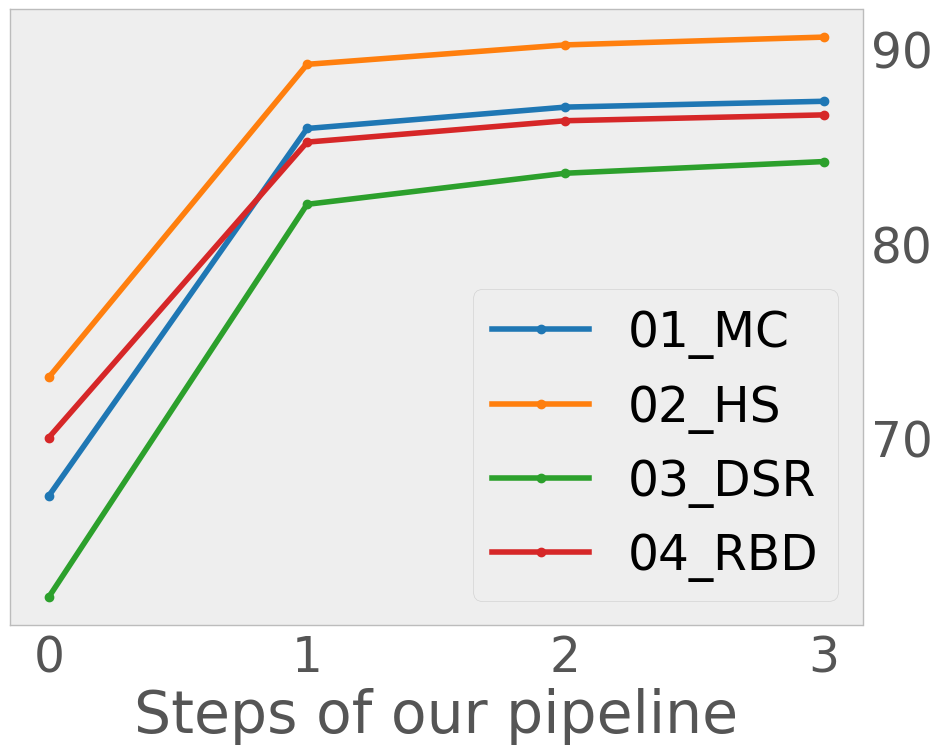}
      \caption{Recall $\uparrow$ of saliency maps}
      \label{fig:}
    \end{subfigure}%

    \begin{subfigure}{.50\textwidth}
      \includegraphics[width=1.00\linewidth]{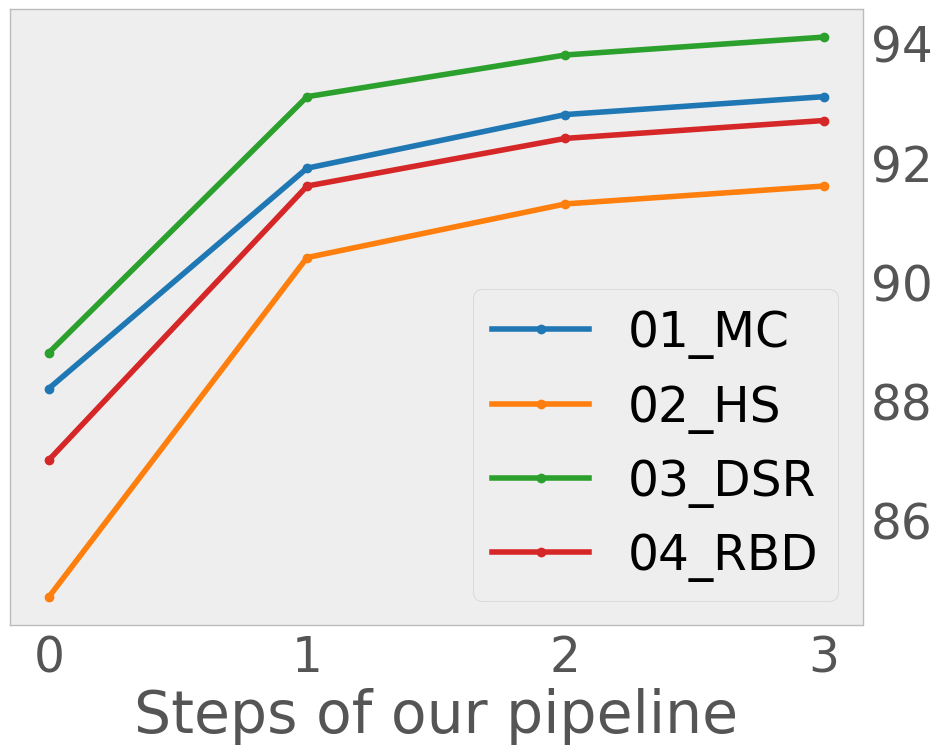}
      \caption{Precision $\uparrow$ of MVA-maps}
      \label{fig:}
    \end{subfigure}%
    \begin{subfigure}{.50\textwidth}
      \includegraphics[width=1.00\linewidth]{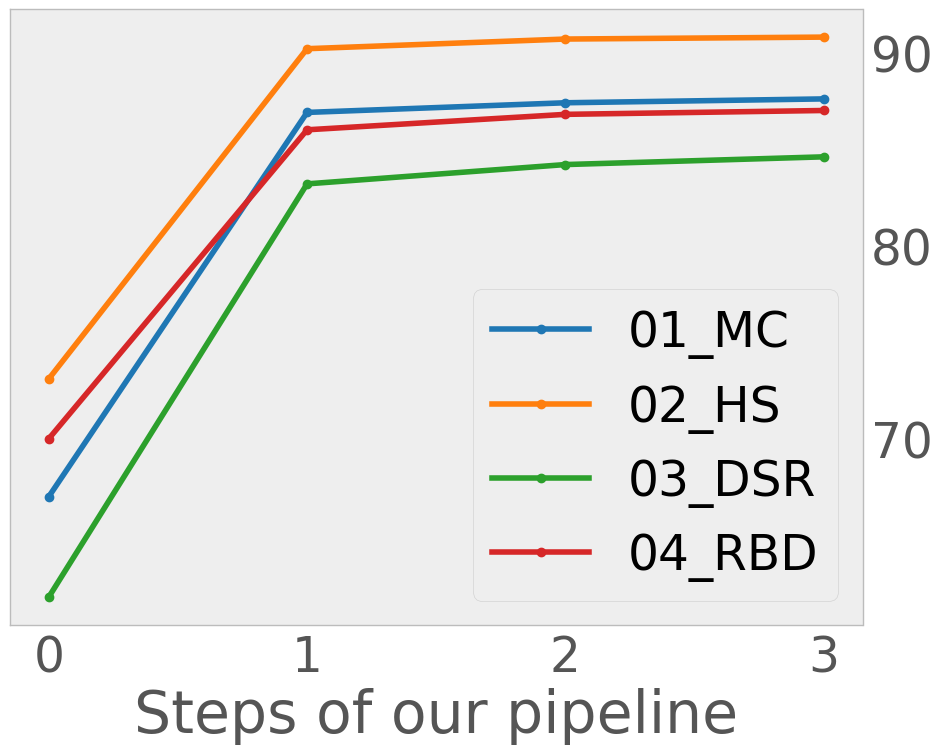}
      \caption{Recall $\uparrow$ of MVA-maps}
      \label{fig:}
    \end{subfigure}%

    \caption{Illustrating the pseudo labels quality improvement by inter-images-consistency learning and self-supervision using the historical moving averages as new targets on the MSRA-B training set accessed using precision and recall. The scores are measured using the hold-out \emph{ground-truth labels}, for network predicted pseudo labels (saliency output maps), %
    and aggregated MVA maps (historical moving averages). Note that the ground truth labels are only used for measuring the quality of pseudo labels and not used during training. We stop the process of iterative self-supervision when the MVA-maps have stabilized, i.e., the changes in subsequent iterations are negligible. Here, the x-axis labels 0-3 represent measure on pseudo labels obtained at different stages in our pipeline in the following order: pseudo labels of handcrafted methods, inter-images-consistency training, refined pseudo labels from the first iteration of self-supervision and the pseudo labels from the second iteration of self-supervision.
    }
    \label{fig:self-supervision all precision and recall}
\end{figure}

Fig.. \ref{fig:self-supervision all precision and recall} shows how the quality of pseudo-labels for training samples are improved quantitatively. The quality is measured using precision and recall on the training set with respect to the \emph{hold-out ground-truth labels}. Note that these ground-truth labels are not used for any training step. Here, the diversity among different methods can be seen clearly. Some methods are superior in terms of precision but inferior in terms of recall.
\vspace{0.5cm}

\newpage
\section{Failure Cases}
\begin{figure}[h]
\centering
        \includegraphics[width=0.8\linewidth]{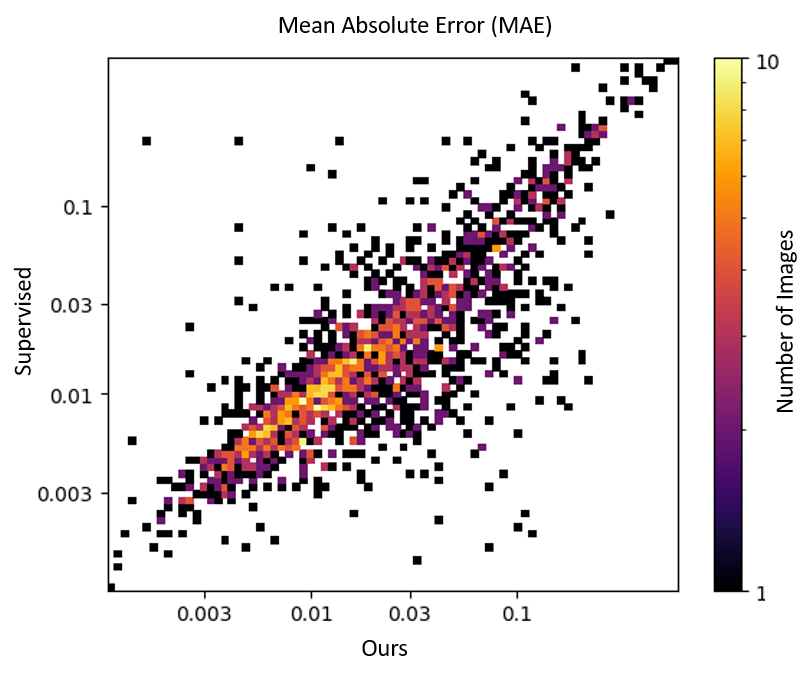}
        \caption{Comparison of the MAE scores of our predictions (x-axis) and those of the baseline from the supervised setting (y-axis). The data indicates a strong correlation of the quality between the predictions of both settings. %
        }
        \label{fig:MAE_Correlation}
\end{figure}

As shown in figure \ref{fig:MAE_Correlation} the correlation between the performance of our unsupervised approach and the supervised (orcale) baseline is strong. In particular, there is a large overlap of failure cases). 
\vspace{0.5cm}

\end{document}